\documentclass[10pt,twocolumn,letterpaper]{article}
\usepackage{times}
\usepackage{epsfig}
\usepackage{graphicx}
\usepackage{amsmath}
\usepackage{amssymb}
\usepackage{amsfonts,amssymb}
\usepackage{makecell}
\usepackage{graphicx}
\usepackage{bbding}
\usepackage{subcaption}
\usepackage{tikz}
\usepackage{multirow}
\usepackage{fontawesome}
\usepackage{latexsym}
\usepackage[numbers,sort&compress]{natbib}

\usepackage[pagenumbers]{iccv} 

\definecolor{iccvblue}{rgb}{0.21,0.49,0.74}
\usepackage[pagebackref,breaklinks,colorlinks,allcolors=iccvblue]{hyperref}

\def\paperID{1932} 
\def\confName{ICCV}
\def\confYear{2025}

\title{\includegraphics[width=0.6cm]{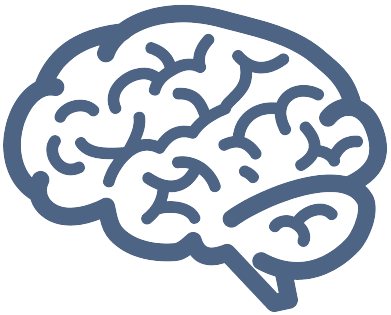} MemorySAM: Memorize Modalities and Semantics with Segment Anything Model 2 for Multi-modal Semantic Segmentation}

\author{Chenfei Liao$^{1}$\quad Xu Zheng$^{1,3,}$\thanks{Project lead}\quad Yuanhuiyi Lyu$^{1}$\quad Haiwei Xue$^{4}$\quad Yihong Cao$^{5}$\\ Jiawen Wang$^{6}$\quad Kailun Yang$^{5}$\quad Xuming Hu$^{1,2,}$\thanks{Corresponding author.} \\ 
$^{1}$AI Thrust, HKUST(GZ) \quad $^{2}$CSE, HKUST \quad $^{3}$INSAIT \quad $^{4}$Tsinghua \quad $^{5}$HNU \quad$^{6}$CUMTB \\
}

\begin{document}

\maketitle
\begin{abstract}
Research has focused on Multi-Modal Semantic Segmentation (MMSS), where pixel-wise predictions are derived from multiple visual modalities captured by diverse sensors. Recently, the large vision model, Segment Anything Model 2 (SAM2), has shown strong zero-shot segmentation performance on both images and videos. When extending SAM2 to MMSS, two issues arise: \textcircled{1} How can SAM2 be adapted to multi-modal data? \textcircled{2} How can SAM2 better understand semantics? 
Inspired by cross-frame correlation in videos, we propose to treat multi-modal data as a sequence of frames representing the same scene. 
Our key idea is to \textbf{``memorize''} the modality-agnostic information and \textbf{``memorize''} the semantics related to the targeted scene.
To achieve this, we apply SAM2’s memory mechanisms across multi-modal data to capture modality-agnostic features. 
Meanwhile, to memorize the semantic knowledge, 
we propose a training-only Semantic Prototype Memory Module (SPMM) to store category-level prototypes across training for facilitating SAM2’s transition from instance to semantic segmentation.
A prototypical adaptation loss is imposed between global and local prototypes iteratively to align and refine SAM2's semantic understanding. 
Extensive experimental results demonstrate that our proposed MemorySAM outperforms SoTA methods by large margins on both synthetic and real-world benchmarks \textbf{(65.38\% on DELIVER, 52.88\% on MCubeS)}. 
Source code is available at \href{https://github.com/Chenfei-Liao/MemorySAM}{https://github.com/Chenfei-Liao/MemorySAM}.
\end{abstract}

\section{Introduction}
\label{sec1}

\begin{figure}[ht]
\centering

\begin{subfigure}{\linewidth}
\includegraphics[width=\textwidth]{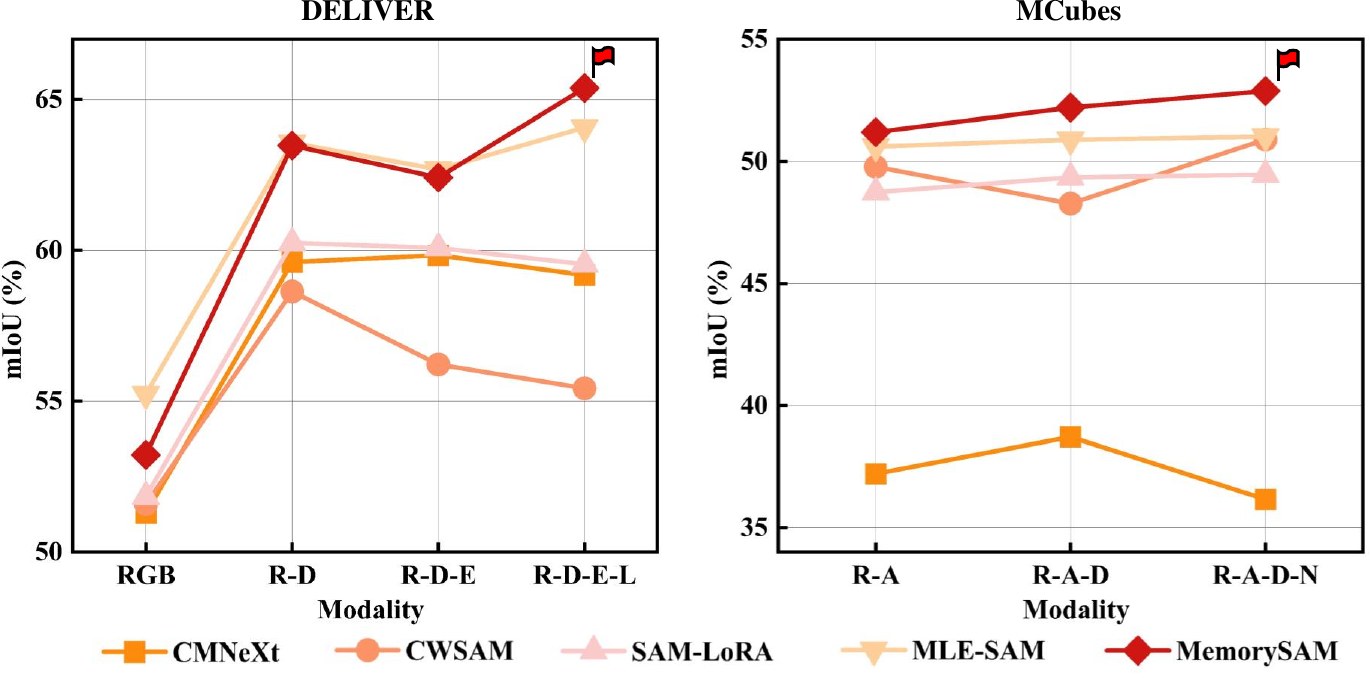} 
\caption{}
\label{fig:memorysam}
\end{subfigure}

\vspace{0.1cm} 

\begin{subfigure}{\linewidth}
\includegraphics[width=\textwidth]{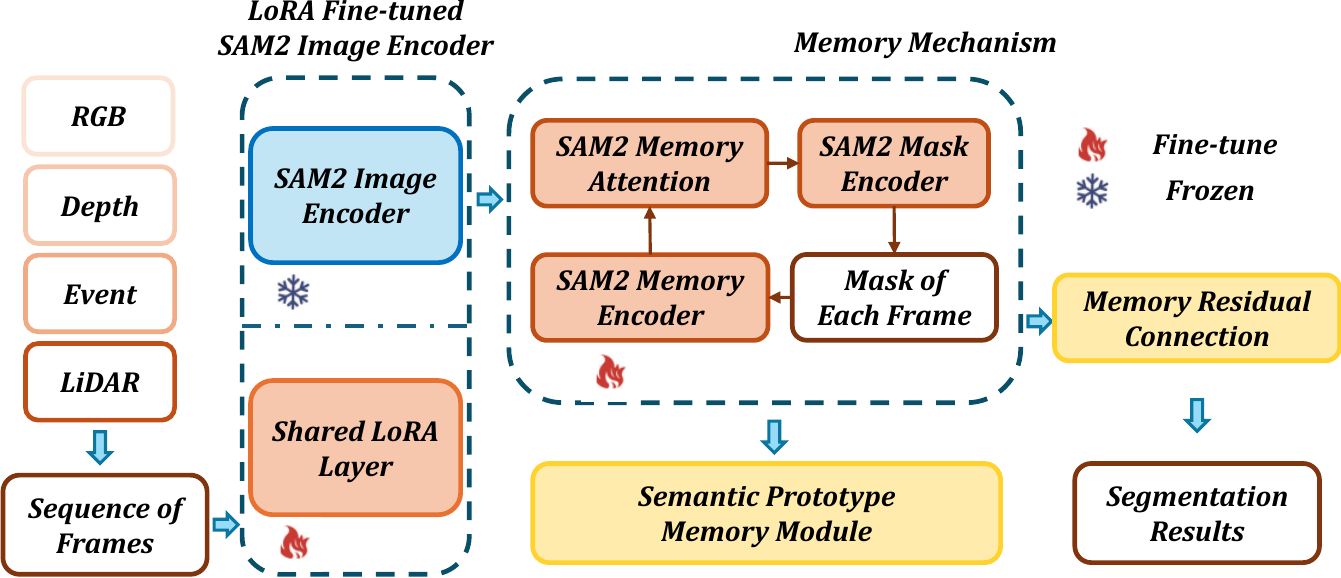} 
\caption{}
\label{fig:performance}
\end{subfigure}
\vspace{-15pt}
\caption{(a) Performance comparison on DELIVER (RGB-Depth-Event-LiDAR Modalities), (b) Overall of MemorySAM.}
\label{fig:teaser}
\end{figure}

In real-world applications, uni-modal data often faces significant challenges in extreme environments. For instance, during nighttime driving~\cite{liu2023improving}, low light conditions~\cite{liu2021benchmarking, wang2024real} and motion blur~\cite{zhang2024generative} cause RGB cameras to lose crucial details, resulting in degraded image quality. To address these limitations, it is essential to incorporate more visual modalities, such as depth data~\cite{cao2021shapeconv, seichter2021efficient}, event data~\cite{zheng2023deep, zhou2024exact} and so on. These complementary modalities can significantly enhance the robustness and accuracy of segmentation models, especially in challenging scenarios~\cite{zhu2024customize, zheng2024magic++,zheng2024learning}.

Meanwhile, the Segment Anything Model 2 (SAM2) has brought a new breakthrough in instance segmentation and obtained an impressive capability in zero-shot scenarios for both images and videos~\cite{ravi2024sam}. SAM2 incorporates a memory mechanism to capture the relationship between frames, enabling it to process video data effectively.
When adapting SAM2 to the Multi-Modal Semantic Segmentation (MMSS), two intuitive challenges occur, including 
\textcircled{1} \textbf{\textit{Adapting SAM2 to Multi-Modal Data:}}
Each modality has distinct characteristics, while SAM2's pre-trained model is tailored for RGB data. 
\textcircled{2} \textbf{\textit{Enhancing Semantic Understanding:}}
Although SAM2 excels in instance segmentation, it exhibits limitations in acquiring and leveraging semantic knowledge for semantic segmentation tasks.

To address these challenges, we propose a novel \textbf{\textit{MemorySAM}} framework that utilizes a new multi-modal processing paradigm and applies semantic understanding modules and reutilizes the memory mechanisms of the SAM2 model.
Our key idea is to ``\textbf{\textit{memorize}}'' the modality-agnostic information and ``\textbf{\textit{memorize}}'' the semantics related to the targeted scene.
The modality-agnostic information focuses on the global structure and common features of a scene, while scene-specific semantics emphasizes detailed object categories and attributes.
For challenge \textcircled{1}, we first fine-tune SAM2’s image encoder to enable SAM2’s adaptation to multi-modal data.
Then, different from the recent state-of-the-art (SoTA) methods, which input the paired data simultaneously, MemorySAM considers the multi-modal data as a \textbf{\textit{sequence}} of frames, applying the memory mechanism of SAM2 to capture the cross-modal modality-agnostic features. 
Concretely, the current modality in the multi-modal sequence integrates previous modalities' features from the memory encoder. 
By doing so, MemorySAM effectively overcomes the challenge and captures modality-agnostic features.

For challenge \textcircled{2}, we propose a Semantic Prototype Memory Module (\textbf{SPMM}) to help SAM2 store semantic knowledge. 
We construct category-level semantic prototypes to encapsulate class-specific knowledge during iteration. 
In the image encoder features, only the class-related pixels are conveyed to the corresponding prototypes, while other pixels are masked. 
During iteration, there exist two types of prototypes: the current prototype and the global prototype. The current prototype is derived from the features of the current iteration. The global prototype is updated based on the current prototype using momentum. By bringing the two prototypes closer via the prototypical knowledge adaptation loss, the semantic knowledge is injected into SAM2, thereby enhancing SAM2's ability for semantic understanding. Through this \textbf{training-only} SPMM, semantic knowledge is brought to SAM2 during the iteration without adding learnable parameters.

Extensive experiments on two public datasets demonstrate that MemorySAM outperforms SoTA methods~\cite{zhang2023delivering, pu2024classwise} for MMSS.
Overall, our contributions are as follows:
\textbf{(I)} We propose a new paradigm to deal with multi-modal data by treating paired multi-modal data as a sequence of frames representing the same scene. 
\textbf{(II)} We propose a novel framework, namely MemorySAM, the first to apply SAM2's memory mechanism in multi-modal semantic segmentation.
\textbf{(III)} To enhance SAM2's ability for semantic understanding, we propose a training-only semantic knowledge memory module, consisting of the prototype memory module and the prototypical knowledge adaptation loss. 
\textbf{(IV)} We conduct extensive experiments on 2 commonly considered datasets, including both real-world and synthetic scenarios. MemorySAM excels most SoTA methods by large margins, reaching 65.38\% on DELIVER~\cite{zhang2023delivering} and 52.88\% on MCubeS~\cite{liang2022multimodal}).

\section{Related Works}
\subsection{Segment Anything Model (SAM) Family}
SAM~\cite{kirillov2023segment} and SAM2~\cite{ravi2024sam} bring a surprising advancement to the instance segmentation task. Different from SAM, SAM2 introduces a memory mechanism to capture the relationship between frames, equipping the model with the capability to deal with videos. Current works based on SAMs can be divided into two directions: improved frameworks for SAMs~\cite{hu2024relax, li2024sam, yang2024samurai, ding2024sam2long} and applications of SAMs~\cite{ wu2023medical, qin2024db, ma2024segment, fu2024cosam, yan2023ringmo, liu2024pointsam, liu2024rsps, huo2024sam,  li2024fusionsam, ma2024manet, xiao2024segment}.

For the former, researchers attempt to optimize SAMs based on their existing limitations. Particularly, AM-SAM~\cite{li2024sam} designs an automated prompting strategy to enhance the model’s segmentation performance and accelerate convergence during training. 
For the latter, researchers attempt to optimize SAMs for specific scenarios and tasks. When applying SAMs to medical images~\cite{ wu2023medical, qin2024db, ma2024segment, fu2024cosam}, remote sensing images~\cite{yan2023ringmo, liu2024pointsam, liu2024rsps}, \textit{etc.}, Adapters~\cite{houlsby2019parameter, chen2023sam} and LoRA~\cite{hu2021lora} are applied to improve SAMs' ability in a specific domain, followed by some targeted modules. Moreover, when applying SAMs to other tasks like image translation~\cite{huo2024sam}, multi-modal semantic segmentation~\cite{ma2024manet, li2024fusionsam, xiao2024segment}, \textit{etc.}, some novel frameworks are proposed to deal with the task differences. However, they only treat the SAMs as components of the whole framework, overlooking the specific characteristics of SAM2 and resulting in an inevitable loss of its full potential. 

In contrast, our approach seeks to explore SAM2's distinct capabilities for multi-modal semantic segmentation (MMSS) rather than simply utilizing it as is. From a multi-modal perspective, we conceptualize the multi-modal data as a sequence of frames representing the same scene, which aligns seamlessly with SAM2's memory mechanism. From a semantic perspective, we introduce a semantic prototype memory module that requires no additional trainable parameters, focusing on enhancing SAM2's ability to comprehend and utilize semantic information.

\begin{figure*}[t!] 
    \centering 
    \includegraphics[width=0.90\textwidth]{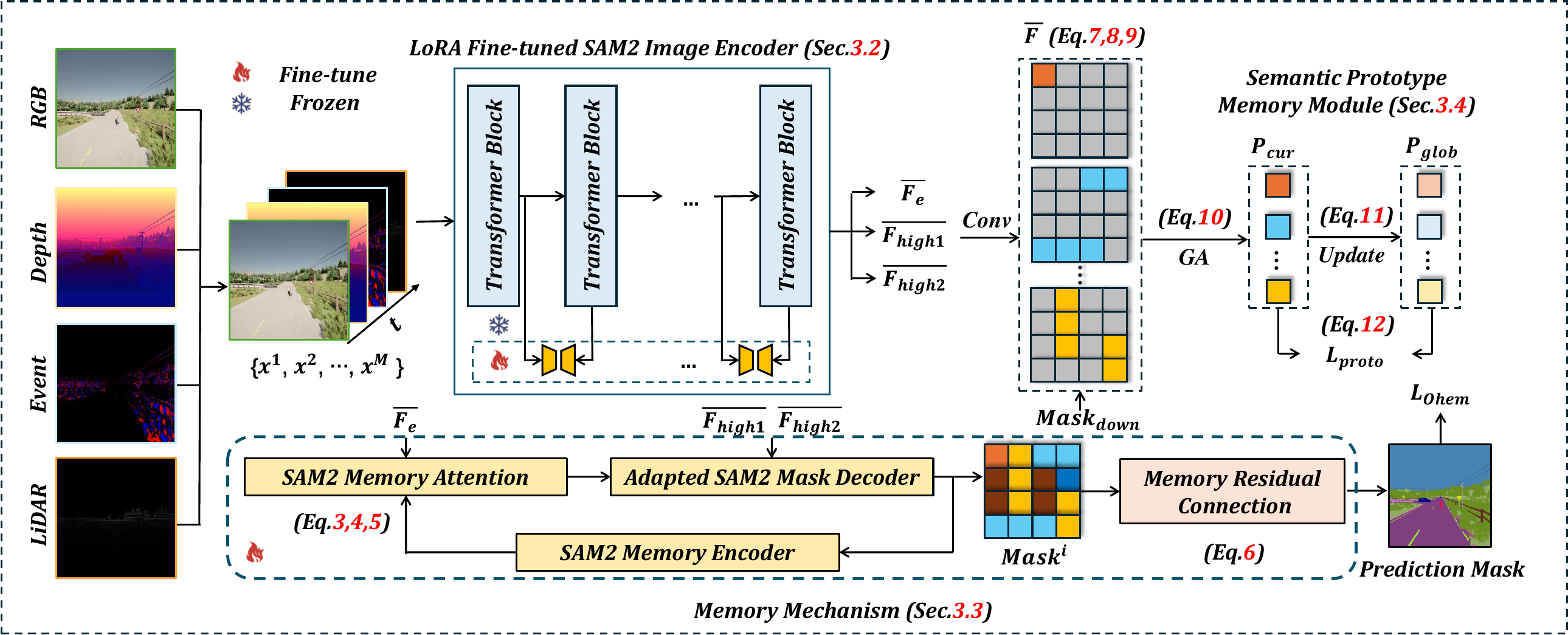}
    \caption{Overall framework of the proposed MemorySAM, it innovatively treats the multi-modal data as sequences and aims to \textbf{``memorize''} the modality-agnostic information and \textbf{``memorize''} the semantics related to the targeted scene.}
    \label{fig:example} 
\end{figure*}

\subsection{Multi-modal Semantic Segmentation}
Semantic segmentation is a key task in computer vision where the goal is to classify every pixel in an image~\cite{mo2022review}. Typically, most related works apply an encoder-decoder architecture based on RGB data for semantic segmentation~\cite{segformer, isnet, enet, fastscnn, bisenet, bisenetv2, stdc, pp-liteseg, rtformer, ddrnet}. However, considering the limited information of unimodal data and the requirements for robustness in practical applications, data of other modalities is introduced, such as the depth~\cite{chen2020bi, cao2021shapeconv, chen2021spatial, seichter2021efficient, yin2023dformer}, thermal~\cite{deng2021feanet, li2022rgb, wu2022complementarity, zhou2023dbcnet, zhao2023mitigating, zhao2025unveiling}, LiDAR~\cite{el2019rgb, borse2023x}, and event data~\cite{zhang2021issafe, xie2024eisnet, kachole2023bimodal, yao2024sam}. Recently, due to the improvement of multi-modal datasets and the development of novel cameras, the focus of related research has gradually shifted from bimodality to multi-modality, such as MCubeSNet~\cite{liang2022multimodal}, CMX~\cite{zhang2023cmx}, CMNeXt~\cite{zhang2023delivering}, and MAGIC~\cite{zheng2024centering}.

Under the premise that multi-modal data is input in pairs, the existing multi-modal semantic segmentation methods can be divided into three types: RGB-centered, RGB-uncentered, and X-centered. RGB-centered methods make the RGB modality the center and the other modalities the auxiliary~\cite{zhang2023cmx, zhang2023delivering, man2023bev}. RGB-uncentered methods treat modalities equally and use a symmetrical structure to fuse multi-modal features~\cite{zhang2021abmdrnet, zhou2022frnet, xiao2022attribute, zhang2023frame, xie2024eisnet, cao2024chasing}. X-centered methods determine the primary modality by measuring the contribution of each modality~\cite{zheng2024centering}. Particularly, MAGIC~\cite{zheng2024centering} ranks the multi-modal features with their similarity with the aggregated features, achieving semantic segmentation with higher accuracy and more robustness.
Differently, in our work, we attempt to tune large vision models for MMSS. Moreover, we rethink the data utilization of multi-modal data from the perspective of sequences, which significantly differs from CMNeXt (RGB-Centered) and MAGIC (X-Centered).

\section{Method}
\label{Method}
In this paper, we strive to address the multi-modal semantic segmentation (MMSS) task with the compelling SAM2 model and introduce a novel MemorySAM framework. 
As illustrated in Figure~\ref{fig:example}, MemorySAM consists of the LoRA fine-tuned SAM2 image encoder, the memory mechanism, and the semantic prototype memory module (SPMM).

Firstly, we consider the paired multi-modal data as a sequence of frames, making them go through the entire framework in order. 
The whole framework begins with the fine-tuning of SAM2's image encoder (Section~\ref{Finetune}). 
Then, the original SAM2's memory mechanism is adapted to MMSS (Section~\ref{Memory}). The core insight of MemorySAM is to consider the multi-modal data as a sequence of frames. For each frame, the features go through the memory attention to learn the modality knowledge from the previous frames before conducting the decode operation. 
Thirdly, due to the knowledge gap between the instance segmentation ability of SAM2 and the targeted MMSS task, we introduce a semantic knowledge memory module based on prototype learning methods between the image encoder and the mask decoder (Section~\ref{SKMM}). This effectively stores the semantic knowledge and hugely enhances MemorySAM's capability of understanding semantics.

\subsection{SAM2's Architecture}
\begin{figure}[t!] 
    \centering 
    \includegraphics[width=0.35\textwidth]{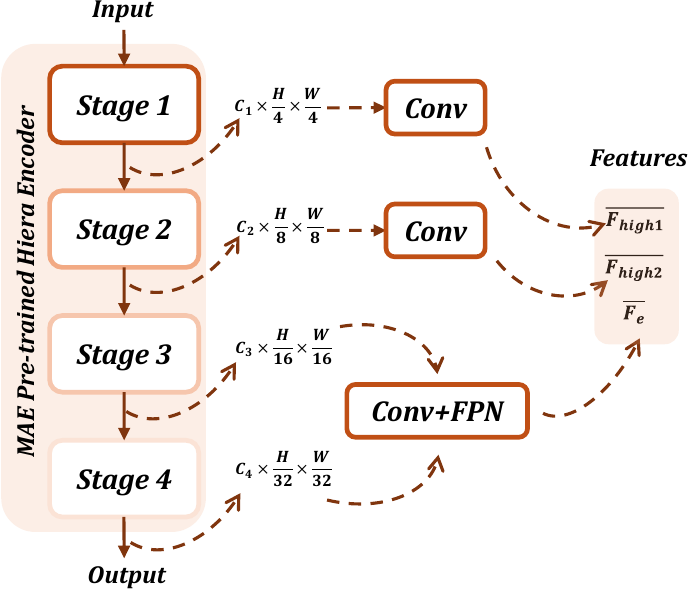}
    \vspace{-4pt}
    \caption{The MAE pre-trained Hiera encoder structure.}
    \label{fig:hiera} 
    \vspace{-8pt}
\end{figure}

SAM2 applies the MAE pre-trained Hiera~\cite{ryali2023hiera, bolya2023window} as the image encoder, progressively reducing spatial resolution and providing multi-scale features of each stage. 
Given the resolution of the input image as $H \times W$, the resolution of the $ith$ stage's feature map is $\frac{H}{2^{i+1}} \times \frac{W}{2^{i+1}}$, where $i \in [1, n]$. 
During SAM2's encoding process, 3 types of features are produced. 
Then, as shown in Figure~\ref{fig:hiera}, the high-level features $\overline{F_{e}}=\{F_{e}^1, ... F_{e}^i\}_{i=1}^M$ are obtained by the combination of convolutional layer and the feature pyramid network (FPN) from the fusion of Stage 3 \& 4' features. Meanwhile, the high-resolution features $\overline{F_{high1}}$ and $\overline{F_{high2}}$ are obtained with convolutional layers.
Notably, only the fused features $\overline{F_{e}}$ go through the memory attention while $\overline{F_{high1}}$ and $\overline{F_{high2}}$ are passed to the mask decoder for providing high-resolution details.

SAM2 proposes a memory mechanism to capture the relationship between frames, including a memory encoder module and a memory attention module. The memory encoder is to produce the memory features of each frame while the memory attention is to integrate current features with previous frames. 
Intuitively, MemorySAM directly re-utilizes the SAM2's memory mechanism of SAM2 for multi-modal input data. 
As to the mask decoder, MemorySAM adapts the number of the output classes to the targeted scene and does not introduce additional parameters.

\subsection{Fine-tuning SAM2's Image Encoder}
\label{Finetune}
Given the input multi-modal data from $M$ modalities as $x = \{ x^m \in \mathbb{R}^{C\times H \times W} | m \in [1, M] \}$ , where $C$, $H$, $W$ represent the number of channels, height, width. Moreover, $M$ stands for the total modality number of the input data and $m$ represents a specific modality such as RGB. Initially, the input batch $x$ is reformulated to a sequence of frames as $\overline{x} = \{x^1, x^2, \cdots, x^M \in \mathbb{R}^{C\times H \times W} \}$, which means the multi-modal data sequence goes through the framework one by one in order rather than in pairs.

Given the substantial number of trainable parameters in SAM2's image encoder, and with the goal of preserving its pre-trained knowledge, we utilize the LoRA technique for fine-tuning.
In more detail, we use a LoRA layer to update the Queue and Value metrics. As shown in Eq.~\ref{Equa6}, the LoRA layer is made up of several low-rank matrices: $W_a^Q \in \mathbb{R}^{d_Q \times r}, W_a^V \in \mathbb{R}^{d_V\times r}, W_b^Q \in \mathbb{R}^{r\times d_Q}$, and $W_b^V \in \mathbb{R}^{r\times d_V}$, where $d_Q \gg r$ and $d_V \gg r$.
\begin{equation}
\setlength{\abovedisplayskip}{3pt}
\setlength{\belowdisplayskip}{3pt}
    \Delta Q = W_a^QW_b^Q,  \Delta V = W_a^VW_b^V.
\label{Equa6}
\end{equation}
The $\Delta Q$ and $\Delta V$ are added to the original Queue metrics as the Eq.~\ref{Equa7}, allowing the effective adaptation of window-based multi-head self-attention with less training cost. Notably, while training, the data from $M$ modalities shares the same LoRA layer, with the backbone parameters frozen.
\begin{equation}
\setlength{\abovedisplayskip}{3pt}
\setlength{\belowdisplayskip}{3pt}
    Q' = Q + \Delta Q,  V' = V + \Delta V.
\label{Equa7}
\end{equation}

\subsection{Memory Mechanism}
\label{Memory}

\begin{figure}[t!] 
    \centering 
    \includegraphics[width=0.42\textwidth]{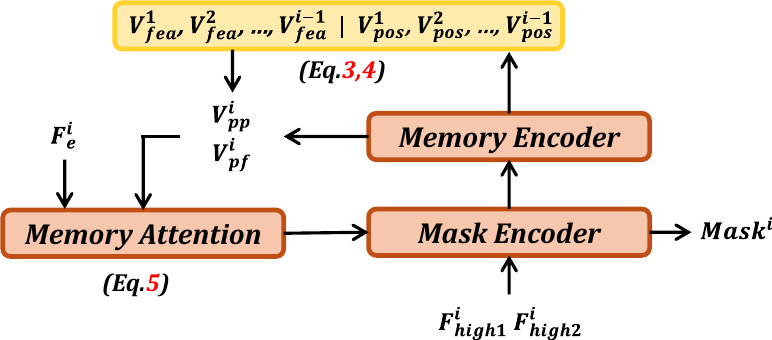}
    \vspace{-4pt}
    \caption{Details of the memory mechanism.}
    \label{fig:memory} 
    \vspace{-8pt}
\end{figure}

For the memory part, we begin by unfreezing the mask decoder, the memory encoder, and the memory attention during training. 
Notably, \( F_{e}^1 \) directly undergoes the mask encoding operation without passing through the memory attention, while the remaining fused features \( F_{e}^i \) (for \( i \in [2, M] \)) first pass through the memory attention before entering the mask decoder.
For each feature \( F_{e}^i \), we calculate two auxiliary variables: the \( i \)-th previous feature, \( V_{pf}^{i} \), and the \( i \)-th previous positional encoding, \( V_{pp}^{i} \), using the following equations:
\begin{equation}
      V_{pf}^i = \text{Concat}(V_{fea}^{1}, V_{fea}^{2}, \dots, V_{fea}^{i-1}),  
\end{equation}
\begin{equation}
    V_{pp}^i = \text{Concat}(V_{pos}^{1}, V_{pos}^{2}, \dots, V_{pos}^{i-1}).
\end{equation}
Once the features and positional embeddings are obtained, the output of the memory encoder—composed of \( V_{pf}^{i} \) and \( V_{pp}^{i} \)—is passed to the memory attention.
The memory attention operation, denoted as \( Att_m \), computes the features from each specific modality using the memory features from previous frames. This process allows the extraction of modality-agnostic, scene-targeted features across different modalities, as shown in the following equation:
\begin{equation}
    F_{c}^i = Att_m(F_{e}^i, (V_{pf}^{i}, V_{pp}^{i})).
\end{equation}
We define the resulting features set obtained from memory attention as \( \overline{F_{c}} = \{F_{c}^2, F_{c}^3, \dots, F_{c}^M \in \mathbb{R}^{C_e \times \frac{H}{16} \times \frac{W}{16}} \} \). 
The resulting feature \( F_{c}^i \) is then passed to the mask decoder, which produces the corresponding prediction mask, \( Mask^i \in \mathbb{R}^{1 \times H \times W} \), for \( i \in [1, M] \). For all \( i \in [1, M-1] \), the masks \( Mask^i \) are passed through the memory decoder, which generates the vision features \( V_{fea}^{i} \) and positional encodings \( V_{pos}^{i} \) for the subsequent frame's meomory attention computation. 

The entire process is shown in Figure~\ref{fig:memory}. For instance, if the input is made up of 3 modalities. First, they will all go through the image encoder to form $\overline{F_{e}}$, $\overline{F_{high1}}$, and $\overline{F_{high2}}$. Then, $F_{e}^1$ will directly go through the mask decoder and memory encoder to produce the $Mask^1$, $V_{fea}^1$, and $V_{pos}^1$. For the second modality, $F_{e}^2$ will first conduct the memory attention with $V_{fea}^1$ and $V_{pos}^1$, with the subsequent operations consistent with the first modality. For the third modality, $F_{e}^3$ will first conduct the memory attention with $V_{pf}^3$ and $V_{pp}^3$, which comes from the concatenation of $V_{fea}^1$ and $V_{fea}^2$, $V_{pos}^1$ and $V_{pos}^2$. Considering that the third modality is the last frame, it doesn't need to go through the memory attention.

Moreover, even if the memory mechanism captures the modality-agnostic knowledge between different modalities, the knowledge is distributed to the features of all modalities. Consequently, only choosing the prediction mask of the last modality as the final prediction mask leads to the loss of information. 
To restrain the information loss and ensure both the robustness and accuracy of MemorySAM, we apply the memory residual connection in the production process of the final prediction mask, which is shown in Eq.~\ref{Equa10}. The final prediction mask comes from the average of the prediction masks of all modalities, retaining the extracted valid information as much as possible.
\begin{equation}
\setlength{\abovedisplayskip}{3pt}
\setlength{\belowdisplayskip}{3pt}
    Mask = \frac{1}{M}{\sum\limits^{M}_{i=1}{Mask^i}}.
\label{Equa10}
\end{equation}

\subsection{Semantic Prototype Memory Module (SPMM)}
\label{SKMM}
After memorizing the modality-agnostic scene information, we turn to adapt SAM2 to MMSS, hoping to memorize the semantic knowledge across training.
Thus, we introduce the semantic prototype memory module, which is designed based on prototypical learning and momentum learning. Through building prototypes for each targeted category and memorizing them with $\mathcal{L}_{proto}$, MemorySAM has the ability to understand semantics is further enhanced.

Upon revisiting the instance segmentation and semantic segmentation tasks, the primary distinction between them lies in the ability to understand the semantics of different classes. Therefore, it is natural to propose the semantic segmentation memory module, built on the prototype learning method, to enhance MemorySAM's capacity for semantic understanding.
Prototypical learning is a machine learning method mostly applied in the classification task and unsupervised domain adaptation semantic segmentation task~\cite{yang2018robust, zhou2022rethinking, liu2025class, 10120939, 10741594}. The prototypes are formed first to represent typical instances of different classes, contributing to the classification of new samples. 

In our semantic prototype memory module, we propose two key components: the current prototype and the global prototype. During each iteration, the $Mask \in \mathbb{R}^{1 \times h \times w}$ first are down-sampled to $Mask_{down} \in \mathbb{R}^{1 \times \frac{H}{4} \times \frac{W}{4}}$ to match the height and width of $F_{high1}^i$ from the image encoder. Then, $F^i_{high1}$ goes through a convolution to adjust its channels to 32, producing $F^{'i}_{high1}$. Next, $F_{high1}^{avg}$ is produced as Eq.~\ref{Equabu1} and masked by $Mask_{down}$ to produce the $\overline{F} = \{F^{1}, F^{2}, \cdots F^{c}  \in \mathbb{R}^{32 \times \frac{H}{4} \times \frac{W}{4}} \}$, where $c$ represents the number of classes. For class $t$, the mask value of the $jth$ element of $F^{t}$  is expressed as Eq.~\ref{Equa11}. The calculation of $\overline{F}$ is shown as Eq.~\ref{Equa12}.
\begin{equation}
\setlength{\abovedisplayskip}{3pt}
\setlength{\belowdisplayskip}{3pt}
   F_{high1}^{avg} = \frac{1}{M}{\sum\limits^{M}_{i=1}}{F_{high1}^{'i}},
\label{Equabu1}
\end{equation}
\begin{equation}
\setlength{\abovedisplayskip}{3pt}
\setlength{\belowdisplayskip}{3pt}
    V^{tj}_{mask} = 
    \begin{cases} 
    1, & \text{if } Mask_{down}^{j} = t \\
    0, & \text{if } Mask_{down}^{j} \neq t
    \end{cases},
\label{Equa11}
\end{equation}
\begin{equation}
\setlength{\abovedisplayskip}{3pt}
\setlength{\belowdisplayskip}{3pt}
    F^{t} = F^{avg}_{high1} \cdot V^t_{mask}, t \in [1,c].
\label{Equa12}
\end{equation}

Based on the $\overline{F}$, the global average operation is conducted to produce the current prototype $\overline{P_{cur}} =  \{P^1_{cur}, P^2_{cur}, \cdots, P^c_{cur} \in \mathbb{R}^{32 \times 1 \times 1} \} $, as shown in Eq.~\ref{Equa13}. After each training iteration, each global prototype is updated as Eq.~\ref{Equa14}, where $\mu \in [0, 1]$ is the momentum coefficient. The global prototype stores the semantic knowledge of each class without any trainable parameters. To bring such knowledge to MemorySAM, we impose loss constraints between the global prototype and the current prototype in each training iteration. Thus, we introduce the $\mathcal{L}_{proto}$ to minimize the distance of the two prototypes, as shown in Eq.~\ref{Equa15}, where $p_{glob}$, $p_{cur} \in \mathbb{R}^{1 \times (32 \times c)}$ are the flattened result of $\overline{P_{glob}}$, $\overline{P_{cur}}$.
\begin{equation}
\setlength{\abovedisplayskip}{3pt}
\setlength{\belowdisplayskip}{3pt}
   P^t_{cur} = \frac{1}{M}{\sum\limits^{M}_{i=1}}F^{it}, t \in [1,c],
\label{Equa13}
\end{equation}
\begin{equation}
\setlength{\abovedisplayskip}{3pt}
\setlength{\belowdisplayskip}{3pt}
   P^t_{glob} \leftarrow \mu P^t_{cur} + (1-\mu) P^{t-1}_{glob}, t \in [1,c],
\label{Equa14}
\end{equation}
\begin{equation}
\setlength{\abovedisplayskip}{3pt}
\setlength{\belowdisplayskip}{3pt}
   \mathcal{L}_{proto} = MSE(p_{glob}, p_{cur}) \times \frac{H}{4} \times \frac{W}{4}.
\label{Equa15}
\end{equation}

Except for the $\mathcal{L}_{proto}$, the Online Hard Example Mining Cross-Entropy (OhemCrossEntropy) loss $\mathcal{L}_{Ohem}$~\cite{shrivastava2016training} is introduced, which focuses more on hard-to-predict pixels. Define the ground truth as $GT \in \mathbb{R}^{H\times W}$, where $GT(i,j) \in \{0,1,\cdots,c-1,255\}$ and 255 indicates the ignore label.
The overall loss function is shown as Eq.~\ref{Equa16}, where $\alpha$ is a hyper-parameter to control the relative importance of $\mathcal{L}_{proto}$. Notably, the SPMM only works during training, while other components all work during both training and validation. 
\begin{equation}
   \mathcal{L} = \alpha \cdot \mathcal{L}_{proto} + \mathcal{L}_{Ohem}(GT, Mask).
\label{Equa16}
\end{equation}

\begin{table}[ht]
\centering
\caption{Experimental results on the DELIVER dataset~\cite{zhang2023delivering}, where $\bigstar$  refers to the combination of all modalities.}
\vspace{-8pt}
\setlength{\tabcolsep}{3pt}
\renewcommand{\arraystretch}{1.4}
\resizebox{\linewidth}{!}{
\begin{tabular}{lcccc}
\hline
             \textbf{Method} & \textbf{Modal} & \textbf{Backbone}  & \textbf{mIoU}  & \textbf{$\Delta \uparrow$}        \\ \hline
CMNeXt~\cite{zhang2023delivering}                 & RGB& MiT-B0                   & 51.29            &    -             \\
CWSAM~\cite{pu2024classwise}             & RGB                & ViT-B  w/ Adapter               & 51.59            & +0.30            \\
SAM-LoRA~\cite{zhu2024customize} & RGB                 & ViT-B w/ 1 LoRA &51.84& +0.55\\
MLE-SAM~\cite{zhu2024customize} & RGB                 & Hiera-B+ w/ 4 LoRA & 55.23 &+3.94\\
MemorySAM         & RGB                 & Hiera-B+   w/ 1 LoRA                  &    \textbf{53.22}         &   \textbf{+1.93}             
 \\ 
 
 \hline
 CMNeXt~\cite{zhang2023delivering}                 & R-D                   & MiT-B0                   & 59.61            &    -             \\
CWSAM  ~\cite{pu2024classwise}            & R-D                  & ViT-B w/ Adapter                &58.64            & -0.97            \\
SAM-LoRA~\cite{zhu2024customize} & R-D                 & ViT-B w/ 1 LoRA &60.25& +0.64\\
MLE-SAM~\cite{zhu2024customize} & R-D               & Hiera-B+ w/ 4 LoRA & 63.57 &+3.96\\
MemorySAM         & R-D                  & Hiera-B+   w/ 1 LoRA                  &          \textbf{63.48}   &        \textbf{+3.87}        
 \\ 
 
 \hline
 CMNeXt~\cite{zhang2023delivering}                 & R-D-E                   & MiT-B0                   & 59.84            &    -             \\
CWSAM~\cite{pu2024classwise}              & R-D-E                & ViT-B w/ Adapter                & 56.22            & -3.62            \\
SAM-LoRA~\cite{zhu2024customize} & R-D-E                 & ViT-B w/ 1 LoRA &60.08& +0.24\\
MLE-SAM~\cite{zhu2024customize} & R-D-E               & Hiera-B+ w/ 4 LoRA  & 62.69 &+2.85\\
MemorySAM         & R-D-E                 & Hiera-B+  w/ 1 LoRA                   &   \textbf{62.42}          &      \textbf{+2.58}          
 \\ 
 
 \hline
 CMNeXt~\cite{zhang2023delivering}                 & R-D-E-L                   & MiT-B0                   &59.18            &    -             \\
CWSAM  ~\cite{pu2024classwise}            &  R-D-E-L                & ViT-B w/ Adapter                & 55.43            & -3.75            \\
SAM-LoRA~\cite{zhu2024customize} & R-D-E-L                & ViT-B w/ 1 LoRA &59.54& +0.36\\
MLE-SAM~\cite{zhu2024customize} & R-D-E-L                & Hiera-B+ w/ 4 LoRA & 64.08 &+4.90\\
MemorySAM  $\bigstar$       &  R-D-E-L                & Hiera-B+ w/ 1 LoRA                  &      \textbf{65.38}       &   \textbf{+6.20}             
 \\ 
 
 \hline
\end{tabular}
}
\vspace{-8pt}
\label{DELIVER}
\end{table}

\section{Experiments}

\subsection{Datasets}
\textbf{DELIVER~\cite{zhang2023delivering}:}
DELIVER is a large-scale semantic segmentation dataset designed for multi-modal autonomous driving scenarios. Based on the CARLA simulator~\cite{dosovitskiy2017carla}, the RGB (R), depth (D), event (E) and LiDAR (L) data are collected. DELIVER is made up of 7,885 front-view images with a resolution of $1,042 \times 1,042$. Moreover, the dataset includes cases of cloudy weather, foggy weather, night weather, rainy weather, and sensor failures like motion blur, over-exposure, and so on. The whole dataset is divided into 3 parts with 25 classes: 3,983 for training, 2,005 for validation, and 1,897 for testing. 
\textbf{MCubeS~\cite{liang2022multimodal}:}
MCubeS is a multi-modal dataset for material semantic segmentation. Under challenging outdoor scenarios, 500 image sets are collected. Each of them has a resolution of $1,920 \times 1,080$ and 4 modalities: RGB, near-infrared (NIR), polarization represented by the angle of linear polarization (AoLP) and the degree of linear polarization (DoLP). The whole dataset is divided into 3 parts with 20 classes: 302 for training, 96 for validation, and 102 for testing.

\subsection{Multi-modal Semantic Segmentation Results}

In the comparison experiments, we choose the CMNeXt~\cite{zhang2023delivering}, CWSAM~\cite{pu2024classwise}, SAM-LoRA~\cite{zhu2024customize}, and MLE-SAM~\cite{zhu2024customize} as the main competitors. CMNeXt~\cite{zhang2023delivering} is the most representative and competitive framework for the MMSS task, while CWSAM is one of the most representative and open-source works of extending SAM to the semantic segmentation task.
SAM-LoRA~\cite{zhu2024customize} fine-tunes SAM based on the Low-Rank Adaptation. Given that SAM's training is based on uni-modal data, it would be unfair to directly compare it. Therefore, we use SAM-LoRA for comparison, which has been fine-tuned on the target data.
Moreover, we also introduce MLE-SAM~\cite{zhu2024customize} in the comparison experiments, which uses the entirely same backbone as MemorySAM. However, several extra components are applied in MLE-SAM such as the auxiliary segmentation head, the routers, and so on, hugely increasing the number of parameters. With no extra trainable parameters added, we attempt to explore the performance gaps between MemorySAM and methods like MLE-SAM.
For fairness, we keep the number of parameters of each model similar when evaluating. Thus, MiT-B, ViT-B, and Hiera-B+ separately serve as the backbones of CMNeXt, CWSAM, and MemorySAM. 

\begin{table}[t]
\centering
\caption{Experimental results on the MCubeS dataset~\cite{liang2022multimodal}, where $\bigstar$  refers to the combination of all modalities.}
\vspace{-8pt}
\setlength{\tabcolsep}{3pt}
\renewcommand{\arraystretch}{1.4}

\resizebox{\linewidth}{!}{
\begin{tabular}{lcccc}
\midrule
             \textbf{Method} & \textbf{Modal} & \textbf{Backbone}  & \textbf{mIoU}  & \textbf{$\Delta \uparrow$}        \\ \midrule
CMNeXt~\cite{zhang2023delivering}                 & R-A                   & MiT-B0                   & 37.21            &    -             \\
CWSAM ~\cite{pu2024classwise}             & R-A                 & ViT-B   w/ Adapter               & 49.78          & +12.57           \\
SAM-LoRA~\cite{zhu2024customize} & R-A                 & ViT-B  w/ 1 LoRA  &48.74& +11.53\\
MLE-SAM~\cite{zhu2024customize} & R-A                  & Hiera-B+ w/ 4 LoRA& 50.61 &+13.40\\
 MemorySAM         &  R-A  & Hiera-B+     w/ 1 LoRA                &    \textbf{51.20}         &    \textbf{+13.99}            
 \\ 
 
 \midrule
 CMNeXt~\cite{zhang2023delivering}               &R-A-D                 & MiT-B0                   &38.72           &    -             \\
CWSAM ~\cite{pu2024classwise}             & R-A-D                  & ViT-B   w/ Adapter               &  48.27           & +9.55            \\

SAM-LoRA~\cite{zhu2024customize} & R-A-D                 & ViT-B  w/ 1 LoRA  &49.35& +10.63\\
MLE-SAM~\cite{zhu2024customize} & R-A-D                  & Hiera-B+ w/ 4 LoRA& 50.89 &+12.17\\
MemorySAM         &  R-A-D                  & Hiera-B+  w/ 1 LoRA                   &    \textbf{52.20}         & \textbf{+13.48}               
 \\ 
 
 \midrule
 CMNeXt ~\cite{zhang2023delivering}                & R-A-D-N                  & MiT-B0                   & 36.16           &    -             \\
CWSAM~\cite{pu2024classwise}              &R-A-D-N                   & ViT-B  w/ Adapter                 & 50.59         & +14.43         \\
SAM-LoRA~\cite{zhu2024customize} & R-A-D-N                 & ViT-B  w/ 1 LoRA  &49.46& +13.30\\
MLE-SAM~\cite{zhu2024customize} & R-A-D-N                  & Hiera-B+ w/ 4 LoRA &51.02  &+14.86\\
MemorySAM $\bigstar$        & R-A-D-N               & Hiera-B+  w/ 1 LoRA                  & \textbf{52.88}            &  \textbf{+16.72}              
 \\

 \midrule
\end{tabular}
}

\label{MCubeS}
\end{table}

\begin{figure}[t!] 
    \centering 
    \includegraphics[width=0.48\textwidth]{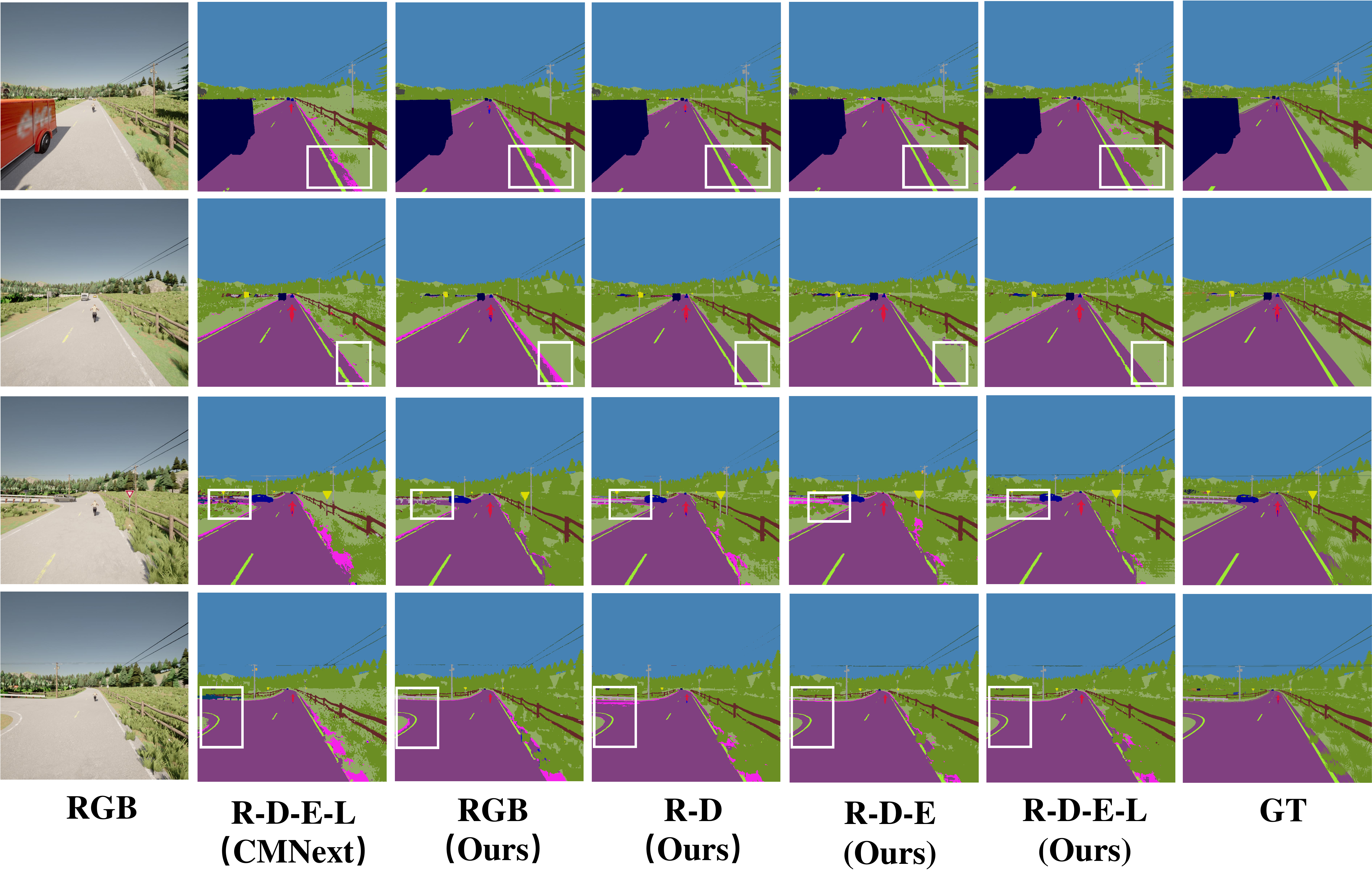}
    \vspace{-20pt}
    \caption{Visual results of MemorySAM on DELIVER (Sun).}
    \label{fig:visual1} 
    \vspace{-4pt}
\end{figure}

\begin{figure}[t!] 
    \centering 
    \includegraphics[width=0.48\textwidth]{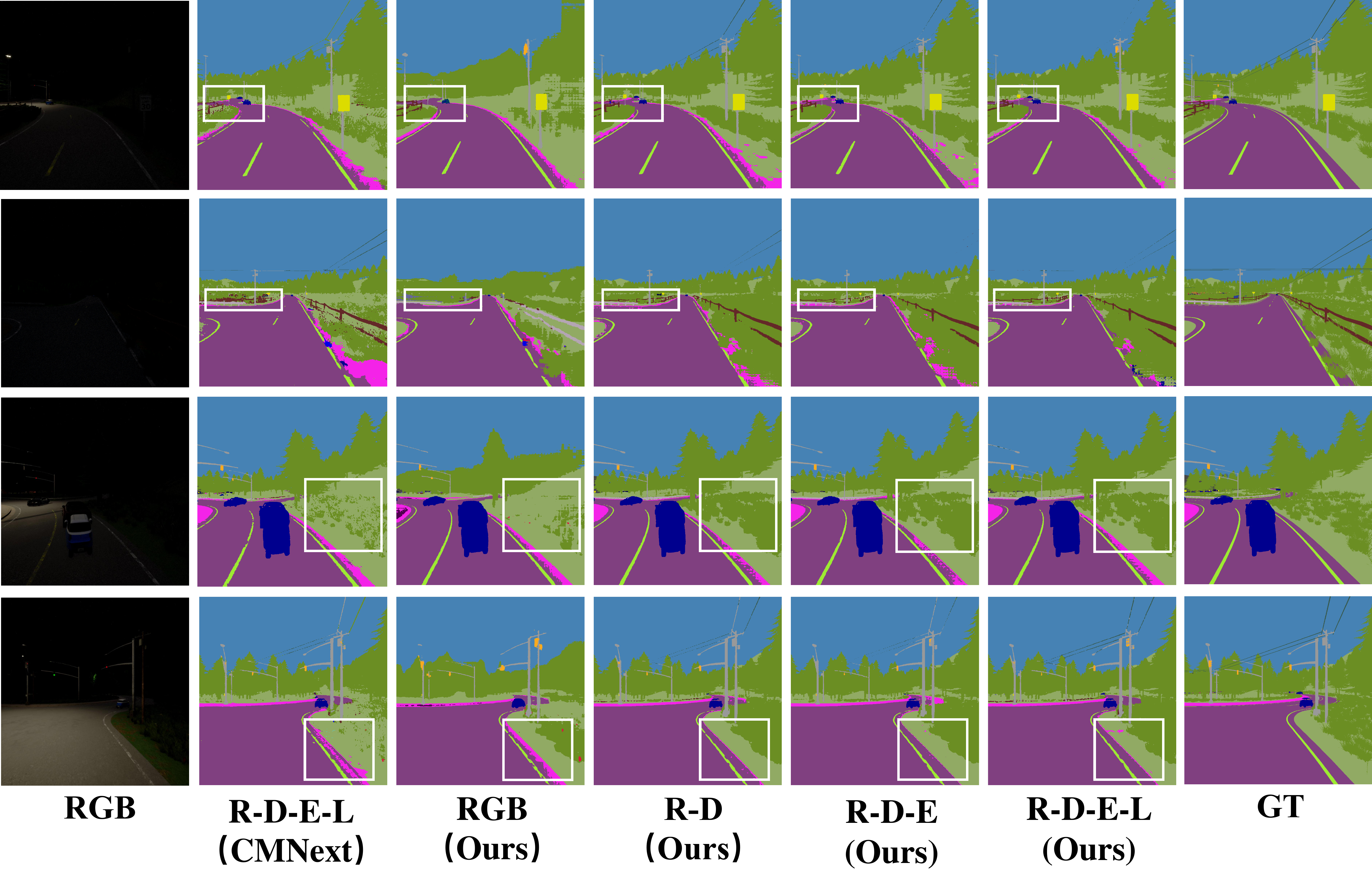}
    \vspace{-20pt}
    \caption{Visual results of MemorySAM on DELIVER (Night).}
    \label{fig:visual2} 
    \vspace{-4pt}
\end{figure}

\noindent \textbf{Results on DELIVER:}
During the comparison experiments on DELIVER, we evaluate the models' performance on RGB, RGB-Depth, RGB-Depth-Event, and RGB-Depth-Event-LiDAR modalities. The results in Table~\ref{DELIVER} demonstrate the efficacy of our proposed MemorySAM. When taking all the modalities as the input, MemorySAM enhances the mIoU by 6.20\% and 9.95\% over CMNeXt and CWSAM. Even when compared with MLE-SAM, MemorySAM still achieves a higher mIoU of 65.38\%. These results show the surprising power of MemorySAM in multi-modal autonomous driving scenarios, proving the effectiveness of our answers to the challenges proposed in Section~\ref{sec1}. Figure~\ref{fig:visual1} and~\ref{fig:visual2} presents the semantic segmentation results of different modality combinations of DELIVER.

\begin{figure}[t!] 
    \centering 
    \includegraphics[width=\linewidth]{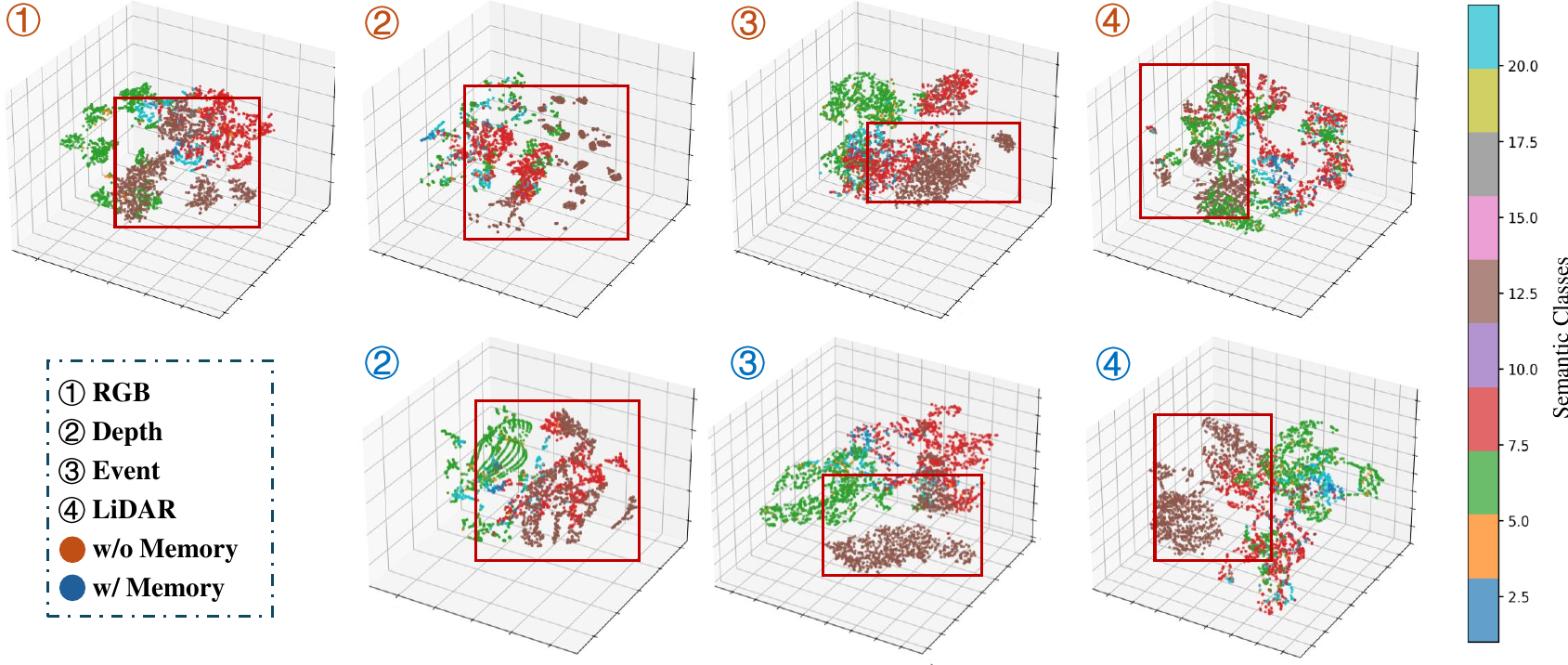}
    \vspace{-12pt}
    \caption{t-SNE results of MemorySAM on DELIVER (RGB-Depth-Event-LiDAR modalities).}
\vspace{-8pt}
    \label{fig:tsne} 
\end{figure}
\noindent \textbf{Results on MCubeS:}
During the comparison experiments on MCubes, we evaluate the models' performance on RGB-AOLP, RGB-AOLP-DOLP, and RGB-AOLP-DOLP-NIR modalities. The results in Table~\ref{MCubeS} further prove the effectiveness of MemorySAM's framework design.  When integrating all of the modalities, MemorySAM attains a mIoU of 52.88\%, surpassing CMNeXt by 16.72\% and MLE-SAM by 1.86\%. The results on MCubeS highlight MemorySAM’s proficiency in multi-modal material semantic segmentation, further exploring MemorySAM's potential.

\subsection{Ablation Study}
\label{Ablation}
Ablation studies are conducted to prove the effectiveness of each component of MemorySAM. Moreover, extra ablation studies are conducted to search for the best momentum of the semantic knowledge memory module and the best weight of $\mathcal{L}_{proto}$. All the experiments in ablation studies are on the DELIVER~\cite{zhang2023delivering} with RGB-Depth modality. 

\begin{table}[t]
\centering
\caption{Ablation study on DELIVER using RGB-D modalities.}
\vspace{-8pt}
\setlength{\tabcolsep}{1pt}
\resizebox{0.75\linewidth}{!}{
\begin{tabular}{ccccc}
\midrule
             \textbf{\makecell{Memory \\ Mechanism}} & \textbf{\makecell{SPMM}} & \textbf{\makecell{Memory\\ Residual Connection}}  & \textbf{mIoU}  & \textbf{ $\Delta \uparrow$ }        \\ \midrule
  \XSolidBrush &  \XSolidBrush &    \Checkmark    &  58.79   & -4.69           \\
  
  \Checkmark         &  \XSolidBrush    &   \XSolidBrush  & 61.70   &  -1.78    \\
  \Checkmark  & \XSolidBrush  &   \Checkmark    & 62.41     &  -1.07\\ 
 \Checkmark &   \Checkmark & \Checkmark  & 63.48 &  -\\
 \bottomrule
\end{tabular}
}
\vspace{-8pt}
\label{Ablation1}
\end{table}

\begin{table}[t]
\centering
\caption{Ablation study on the momentum of SPMM.}
\vspace{-8pt}
\setlength{\tabcolsep}{12pt}
\resizebox{0.75\linewidth}{!}{
\begin{tabular}{cccc}
\midrule
             \textbf{$\alpha_{Proto}$} & \textbf{Momentum} & \textbf{mIoU}    & $\Delta \uparrow$      \\ \midrule
1 & 0.05& 62.54 & -0.94 \\
 1 & 0.1& 62.80 &  -0.68       \\
 1 & 0.2&  63.48 & -       \\
 1 &0.4  &61.47 &   -2.01           \\
 \bottomrule
\end{tabular}
}
\label{Ablation2}
\vspace{-8pt}
\end{table}

\noindent \textbf{Effectiveness of All Modules:}
Table~\ref{Ablation1} evaluates the impacts of the memory mechanism and the SPMM of MemorySAM. And we also evaluate the influence of the memory residual connection in the memory mechanism. 
As shown in Table~\ref{Ablation1}, compared to the SAM2 which only has the shared LoRA layer and the unfrozen memory mechanism, the memory residual connection brings an increase of mIoU by 0.71\%. Compared to the SAM2 which only has the shared LoRA layer and the memory residual connection, the unfrozen memory mechanism brings an increase of mIoU by 3.62\%. Last but not least, the semantic knowledge mechanism achieves another increase of mIoU by 1.07\%. All the performance improvements validate the effectiveness of all the proposed modules and components.

\begin{table}[t]
\centering
\caption{Ablation study on the weight of $Loss_{proto}$.}
\vspace{-8pt}
\setlength{\tabcolsep}{10pt}
\resizebox{0.75\linewidth}{!}{
\begin{tabular}{cccc}
\midrule
             \textbf{$\alpha_{Proto}$} & \textbf{Momentum} & \textbf{mIoU}    & $\Delta \uparrow$      \\ \midrule
0.125 & 0.8& 61.60 & -1.88 \\
 0.25 & 0.8& 63.02 &  -0.46       \\
 1 & 0.8&  63.48 & -       \\
 4 &0.8  &62.31 &   -1.17           \\
 8  & 0.8&  62.98 &   -0.50         \\ 
 16   & 0.8& 63.24 & -0.24 \\ 
 \midrule
\end{tabular}
}
\vspace{-12pt}
\label{Ablation3}
\end{table}

\noindent \textbf{Sensitivity analysis of hyper-parameters:} For the momentum of the semantic knowledge memory module, we conduct ablation study with the values of 0.05, 0.1, 0.2, and 0.4 separately. As shown in Table~\ref{Ablation2}, when the value is over or under 0.2, the mIoU both fall, proving that 0.2 is the best momentum.  For the weight of $Loss_{proto}$, on the premise of ensuring that the difference between $Loss_{proto}$ and $Loss_{Ohem}$ is not too large, we conduct the ablation study with the values of 0.125, 0.25, 1, 4, 8, and 16 separately. As demonstrated in Table~\ref{Ablation3}, we select 1 as the final weight according to the performance.

\noindent \textbf{t-SNE Results:}
Figure~\ref{fig:tsne} presents the t-SNE visualization~\cite{van2008visualizing} of multi-modal features and the fused MemorySAM features and demonstrates the separation and clustering of features across all modalities. As shown in Figure~\ref{fig:tsne}, the memory mechanism allows for more compact clustering of features of the same category and clearer separation of features of different categories, which further proves the effectiveness of the memory mechanism. \textit{More details can be found in the suppl. mat.}.

\subsection{Discussions}
\label{discuss}

\begin{figure}[t!] 
\vspace{-8pt}
    \centering 
    \includegraphics[width=0.48\textwidth]{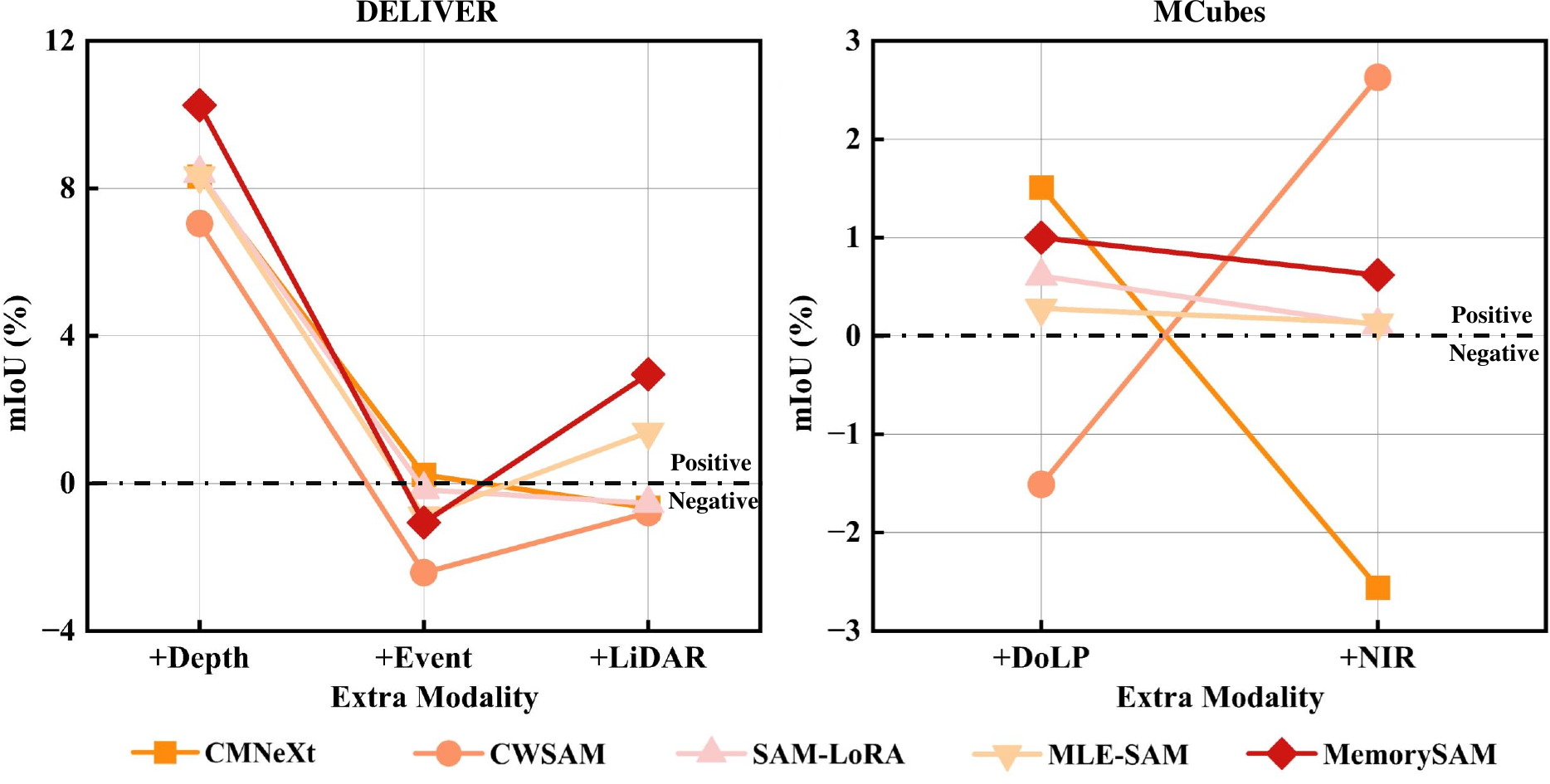}
    \vspace{-20pt}
    \caption{The change in mIoU with each increase in modality.}
\vspace{-8pt}
    \label{fig:delta} 
\end{figure}

\noindent \textbf{Better performance with more modalities?} 
Notably, as shown in Table~\ref{DELIVER} and Figure~\ref{fig:delta}, MemorySAM achieves a significant improvement when adding LiDAR to the input while mIoU of CMNeXt, CWSAM, and SAM-LoRA begin to fall when taking the RGB-Depth-Event-Lidar modality combination as the input, proving MemorySAM's surprising capability of dealing with data which is composed of more than 3 modalities. 
This result further demonstrates our opinion that paired multi-modal data can be considered as a sequence of frames, as discussed in Section~\ref{Memory}. On one hand, as shown in Table~\ref{DELIVER} and~\ref{MCubeS}, except for adding event modality in DELIVER, when the number of modalities increases, the sequence characteristic of multi-modal data is further strengthened. On the other hand, it is beneficial for the memory mechanism to learn modality-agnostic features when faced with more modalities, which is equal to offering the memory mechanism more training samples.

\noindent \textbf{Comparison with MLE-SAM:} 
Moreover, considering that MLE-SAM and MemorySAM have the same backbone, we further calculate the trainable parameters of MLE-SAM and MemorySAM for a quantitative comparison. As shown in Table~\ref{paracompare}, under the RGB-Depth-Event-LiDAR modality combination, the number of trainable parameters of MemorySAM is 2/3 of MLE-SAM's. However, the mIoU of MemorySAM excels MLE-SAM by 1.3\%. When it comes to the RGB-Depth and RGB-Depth-Event modality combinations, the mIoU scores of MemorySAM are so close to those of MLE-SAM, with only 0.09\% and 0.27\% behind respectively. As to Mcubes, Memory-SAM surpasses MLE-SAM under 3 modality combinations, which further proves the effectiveness of our design. Thus, it can be proved that the design of MemorySAM effectively explores the potential of SAM2 in MMSS than the other frameworks.

\begin{table}[t!]
\centering
\caption{Trainable parameters comparison between MLE-SAM and MemorySAM on DELIVER using RGB-D-E-L modalities.}
\vspace{-8pt}
\setlength{\tabcolsep}{6pt}
\renewcommand{\arraystretch}{1.4}
\resizebox{\linewidth}{!}{
\begin{tabular}{ccccc}
\midrule
  \textbf{Model} & \textbf{\makecell{Trainable\\ Parameters}} & $\Delta \uparrow$  &        \textbf{mIoU} & $\Delta \uparrow$ \\ \midrule

MLE-SAM~\cite{zhu2024customize}  &   20.79M & - & 64.08 & -\\ 
 MemorySAM  &   \textbf{14.50M} &6.29M$\downarrow$  &  \textbf{65.38} &\textbf{+1.30}\\ 
\bottomrule
\end{tabular}
}
\label{paracompare}
\vspace{-8pt}
\end{table}

\noindent \textbf{The choice of frozen parts:} In MemorySAM, we directly fine-tune the image decoder, the memory encoder, and the memory attention. For clearer presentation, we conduct several discussion experiments to further validate the rationality by freezing the memory mechanism and the image encoder and the memory mechanism part. 
As shown in Table~\ref{Frozen}, freezing the parameters of both SAM2's image encoder and memory modules leads to significant performance drops.
Thus, it can be proved that the initial weight of SAM2 is not fit for MMSS, coming from the uni-modal training data, \ie, the SA-1B RGB dataset~\cite{ravi2024sam}. Considering that the data characteristics of different modalities are hugely different, it is necessary to fine-tune these 3 components mentioned above to achieve cross-modal adaptation.

\begin{table}[t!]
\centering
\caption{Discussion of the fine-tuned Parts on DELIVER using RGB-D-E-L modalities.}
\vspace{-8pt}
\setlength{\tabcolsep}{10pt}
\renewcommand{\arraystretch}{1.4}
\resizebox{\linewidth}{!}{
\begin{tabular}{cccc}
\midrule
             \textbf{\makecell{SAM2 Image\\ Decoder}} & \textbf{\makecell{SAM2 Memory \\Encoder + Attention}} &   \textbf{mIoU}  & \textbf{ $\Delta \uparrow$ }        \\ \midrule

  \XSolidBrush       &  \XSolidBrush      & 11.65   &  -53.73    \\
   \Checkmark   & \XSolidBrush &   57.00     &  -8.38\\ 
 \Checkmark &   \Checkmark  & \textbf{65.38} &  -\\
 
 \bottomrule

\end{tabular}
}
\vspace{-12pt}
\label{Frozen}
\end{table}
\section{Conclusion}
In this paper, we made the first attempt to deal with multi-modal data from the perspective of videos and apply the memory mechanism to MMSS. We focus on addressing two issues about adapting SAM2 to MMSS: \textcircled{1} How can SAM2 be adapted to multi-modal data? \textcircled{2} How to make SAM2 better understand semantics? Thus, we re-implement the existing memory mechanism of SAM2 to memory multi-modal knowledge and propose the semantic knowledge memory module (SPMM) to memory semantics. Our proposed MemorySAM outstrips previous state-of-the-art MMSS methods on three public multi-modal benchmarks, including both real-world and synthetic scenarios. 
Through the extra experiments in Section~\ref{discuss}, we have some interesting findings: (1) With modality increasing, MemorySAM has better segmentation performance; (2) With proper design, SAM2 can have better performance on MMSS without adding huge extra parameters; (3) Except for the image encoder, the more components of SAM2 fine-tuned, the better performance SAM2 has on MMSS.

\clearpage
{
    \small
    \bibliographystyle{ieeenat_fullname}
    \bibliography{main}
}
\clearpage
\appendix
\clearpage
{\centering
    \textbf{\Large  ==Supplementary Material==} \par
}
\section{More Experimental Details}

\subsection{Dataset Details}

\noindent \textbf{DELIVER \cite{zhang2023delivering}:}
DELIVER is an extensive semantic segmentation dataset specifically designed for multi-modal autonomous driving scenarios. This dataset is built upon the CARLA simulator \cite{dosovitskiy2017carla}, which provides a robust environment for generating diverse driving conditions. The DELIVER dataset encompasses a rich collection of data types, including RGB images (R), depth information (D), event data (E), and LiDAR point cloud data (L). In total, it consists of 7,885 front-view images, each with a resolution of $1,042 \times 1,042$ pixels, ensuring high-quality input for training and evaluation purposes.

Furthermore, DELIVER captures a wide range of environmental conditions that are critical for developing resilient autonomous driving systems. This includes challenging scenarios such as cloudy weather, fog, nighttime driving, and rainy conditions. Additionally, the dataset incorporates various sensor failures, which can significantly impact perception, such as motion blur and over-exposure. These diverse conditions enhance the dataset's applicability in real-world situations.

To facilitate effective training and evaluation, the DELIVER dataset is systematically divided into three distinct subsets: the training set contains 3,983 images, the validation set includes 2,005 images, and the testing set comprises 1,897 images. In total, the dataset is annotated with 25 different classes, providing a comprehensive framework for multi-class semantic segmentation tasks in autonomous driving applications. This structured approach allows researchers and developers to rigorously test and improve their algorithms in a variety of simulated driving environments.

\textbf{MCubeS \cite{liang2022multimodal}:}
MCubeS is a comprehensive multi-modal dataset specifically designed for the task of material semantic segmentation. This dataset was meticulously curated under a variety of challenging outdoor scenarios, resulting in a collection of 500 image sets. Each image set features a high resolution of $1,920 \times 1,080$ pixels, providing detailed visual information essential for accurate segmentation.

The dataset is notable for its inclusion of four distinct modalities, which significantly enrich the data for analysis and model training. These modalities consist of traditional RGB images, near-infrared (NIR) imagery, and polarization data, which is captured in two forms: the angle of linear polarization (AoLP) and the degree of linear polarization (DoLP). This multi-faceted approach allows for a more nuanced understanding of the materials being analyzed, as each modality offers unique insights into their physical properties.

To facilitate effective training, validation, and testing of segmentation algorithms, the entire MCubeS dataset is systematically divided into three subsets. The training set consists of 302 image sets, providing ample data for model development. The validation set includes 96 image sets, which are used to fine-tune model parameters and assess performance during training. Finally, the testing set comprises 102 image sets, serving as an independent benchmark to evaluate the final model's accuracy and effectiveness.

Overall, with its rich multi-modal data and structured division into training, validation, and testing subsets, MCubeS offers a valuable resource for researchers and practitioners aiming to advance the field of material semantic segmentation in complex outdoor environments.

\subsection{Experimental Setups}
Data augmentation methods such as horizontal flipping, Gaussian blurring, random cropping, and so on are applied during the data pre-processing, which is the same as \cite{zhang2023delivering}. The training parameters and configurations for MemorySAM are shown in Table \ref{para1}. The image size for data from DELIVER and MCubes is standardized to 1024 $\times$ 1024, which matches the Segment Anything Model's input size. Considering the limitations of GPUs, the LoRA rank is set to 4. Meanwhile, the batch size is set to 4. During training, the numbers of total training epochs are set to 200 and 500 individually for DELIVER and MCubes. Moreover, the warmup strategy is applied while training, with the number of warmup epochs set to 10.
The learning rates of training processes under different modality combinations and different datasets differs from each other, as shown in \ref{para2}. During our large amounts of experiments, it can be concluded that when the number of modalities turns more, lifting the learning rate slightly may be worth trying.
\begin{table}[ht]
\centering
\caption{Training parameters and configurations for MemorySAM.}
\setlength{\tabcolsep}{12pt}
\renewcommand{\arraystretch}{1.4}
\resizebox{0.9\linewidth}{!}{
\begin{tabular}{ccc}
\hline
             \textbf{ Parameter} & \textbf{Dataset}  &  \textbf{Value}   \\ \hline

  Image Size      &  All & [1024, 1024]       \\ 
  Batch Size    & All & 4 \\ 
    \multirow{2}{*}{Training Epochs} & DELIVER & 200\\
    & MCubes & 500\\
   Optimizer & All &AdamW \\
   Weight Decay & All& 0.01\\
   Scheduler & All & Warmup Poly LR\\
   Scheduler Power & All & 0.9\\
   Warmup  Epochs & All & 10\\
   Warmup Ratio & All & 0.1\\
   LoRA Rank & All & 4 \\
   GPU & All & 4 $\times$ A800\\
   
 \hline

\end{tabular}
}
\label{para1}
\end{table}
\begin{table}[ht]
\centering
\caption{Learning rates for MemorySAM.}
\setlength{\tabcolsep}{12pt}
\renewcommand{\arraystretch}{1.4}
\resizebox{0.9\linewidth}{!}{
\begin{tabular}{ccc}
\hline
   \textbf{Dataset}    &        \textbf{Modality Combination} &  \textbf{Learning Rate}   \\ \hline

  \multirow{4}{*}{DELIVER}   &  RGB &    $4.5 \times 10^{-4} $   \\ 
   &  R-D &   $4.5 \times 10^{-4} $    \\ 
    & R-D-E &   $6 \times 10^{-4} $    \\ 
     & R-D-E-L &  $6 \times 10^{-4} $       \\ 
\hline
     \multirow{3}{*}{MCubes}   &  R-A &   $3.5 \times 10^{-4} $      \\ 
   &  R-A-D &    $4 \times 10^{-4} $      \\ 
    & R-A-D-N &   $3.5 \times 10^{-4} $     \\

 \hline
\end{tabular}
}
\label{para2}
\end{table}

\begin{figure}[htbp] 
    \centering 
    \includegraphics[width=0.5\textwidth]{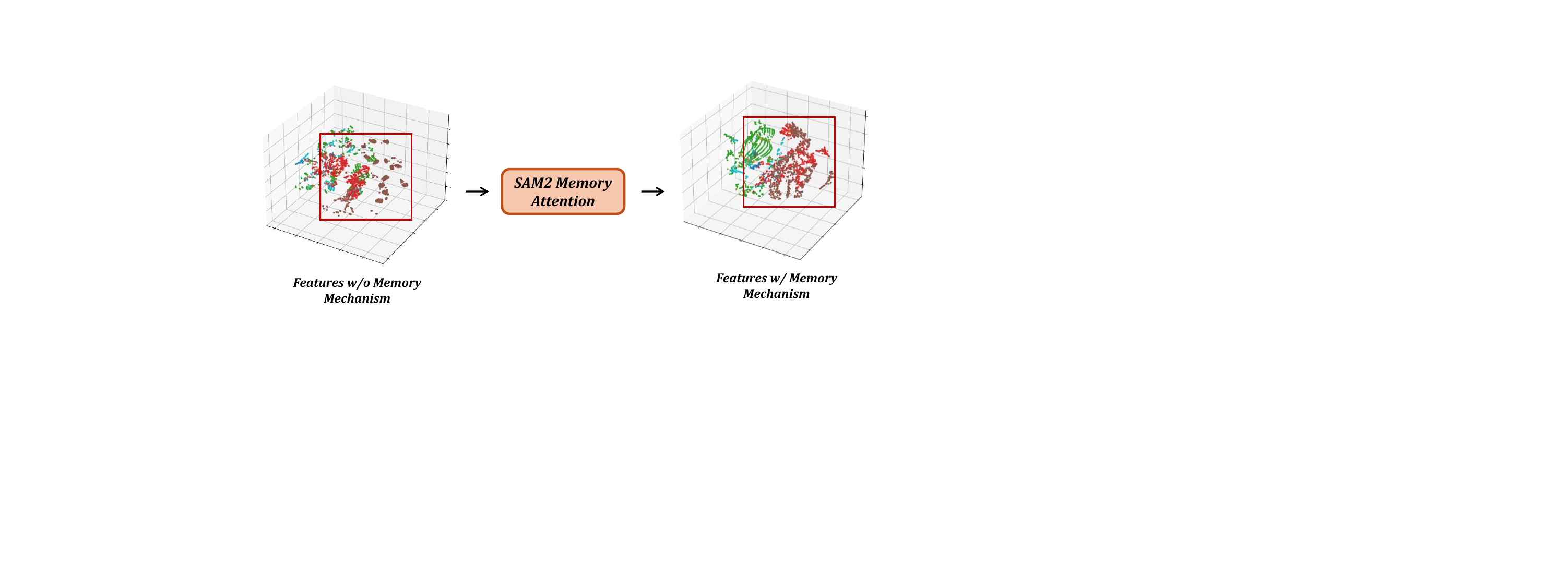}
    \caption{Details of the t-SNE method.}
    \label{fig:detailtsne} 
\end{figure}
\section{Further Elaboration of Experiments Section}

\subsection{Details of t-SNE}
t-SNE~\cite{van2008visualizing} is a classical dimension reduction method. Through t-SNE, the high-dimension features can be transformed into low-dimension data, making it better for data visualization. We use the scikit-learn~\cite{pedregosa2011scikit} to achieve t-SNE, with the parameters set as Table~\ref{t-SNE}, of which most follow the defaults. We apply t-SNE to the features between the image encoder and the mask decoder to validate the influence of the memory mechanism on features. In more detail, we visualize the features before and after the memory mechanism through t-SNE, as shown in Figure~\ref{fig:detailtsne} .

\begin{table}[ht]
\centering
\caption{Parameters of t-SNE.}
\setlength{\tabcolsep}{12pt}
\renewcommand{\arraystretch}{1.4}
\resizebox{0.7\linewidth}{!}{
\begin{tabular}{cc}
\hline
   \textbf{Parameter}    &        \textbf{Value}  \\ \hline

Max Number of Samples  &  5000     \\ 
 Perplexity  &  30     \\ 
  Iteration Times  &   1000     \\ 
  Number of Components & 3 \\
  Random State & 1 \\
  Others  & Defaults \\

 \hline

\end{tabular}
}
\label{t-SNE}
\end{table}

\subsection{More Experiment Results}

\begin{figure}[ht!] 
    \centering 
    \includegraphics[width=0.49\textwidth]{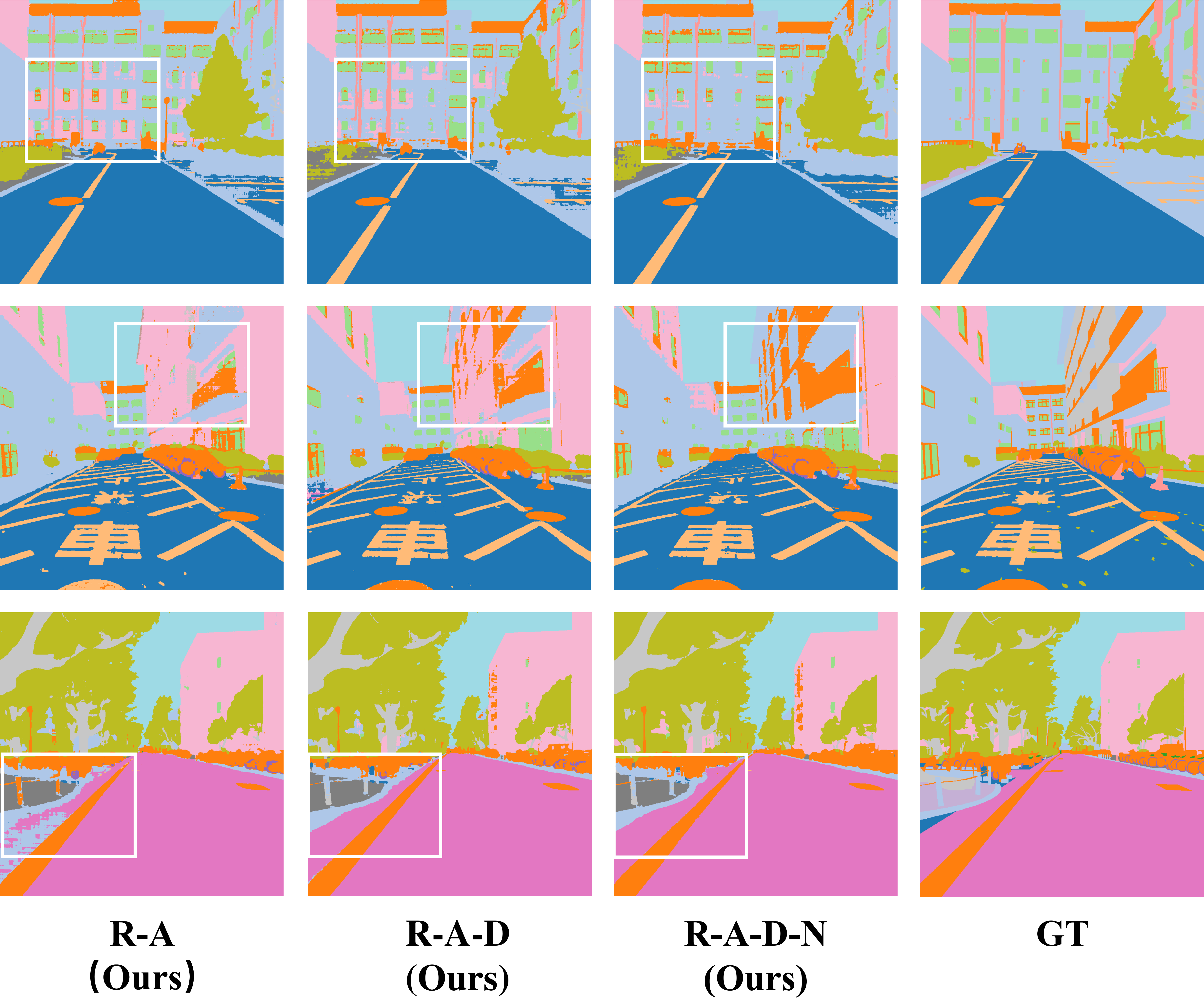}
    \caption{Visual results of MemorySAM on MCubes.}
    \label{fig:mcubes} 
\end{figure}

Due to the limited number of pages, we only show the mIoU in the experimental section. The class-level IoU results are presented in Table \ref{DELIVER_suppl} and \ref{Mcubes_suppl}, which come from the semantic segmentation results under different modality combinations. Red color~refers to the highest IoU of each class while green color~refers to the highest Acc of each class. Moreover, the visualization of MemorySAM on MCubes is shown in Figure \ref{fig:mcubes}.
\clearpage
\begin{table*}[ht]
\centering
\caption{Class-Level semantic segmentation results on DELIVER.}
\setlength{\tabcolsep}{12pt}
\renewcommand{\arraystretch}{1.4}
\resizebox{0.9\linewidth}{!}{
\begin{tabular}{ccccccccc}
\hline
    \multirow{2}{*}{\textbf{Class}}    & \multicolumn{2}{c}{\textbf{RGB}} & \multicolumn{2}{c}{\textbf{R-D}} & \multicolumn{2}{c}{\textbf{R-D-E}} &  \multicolumn{2}{c}{\textbf{R-D-E-L}}   \\
    \cline{2-9} 
    &  \textbf{IoU} & \textbf{Acc} &  \textbf{IoU} & \textbf{Acc} &  \textbf{IoU} & \textbf{Acc} & \textbf{IoU} & \textbf{Acc}  \\ \hline

Building &84.86 & 94.57& 90.21&97.87 & 88.88& 97.94& 89.38 & 98.33     \\ 
 Fence& 27.44& 33.73&36.99 &53.61 &37.63 &59.08  &  41.21 & 53.27     \\ 
 Other&0.00 &0.00 & 0.00&0.00 & 0.00& 0.00 &   0.00 & 0.00     \\ 
  Pedestrian& 65.00& 82.20&74.89 &85.65 &72.54 &86.65& 73.16 & 83.93 \\
 Pole& 60.13& 74.74&73.14 & 85.27& 71.97& 81.55 & 69.45 &82.78 \\
 RoadLine&82.49 &84.91 &84.75 & 88.52&83.59 & 87.61&83.79 &86.81 \\
 Road& 97.83& 99.00& 97.65& 98.51&97.89 &98.64 & 98.11&98.97 \\
 SideWalk&72.05 &92.18 &81.41 & 93.52& 80.47& 92.90& 83.65&93.42  \\
 Vegetation &82.25 &91.73&88.36 &  93.93&88.40 &94.34 & 87.57&92.97 \\
 Cars &89.05 & 96.40&94.31 &97.79 &93.80 & 98.19& 91.74& 98.14 \\
 Wall &55.30 & 68.84& 60.88&78.70 &60.50 &79.27 &56.18 &66.50 \\
 Trafficsign &43.29 &53.32 &72.66 &80.23 &54.87 & 60.94&67.43 &76.83 \\
 Sky &96.44 & 99.32&99.33 &99.73 & 99.26& 99.73& 99.33&99.68 \\
Ground&1.06 & 2.24&1.75 & 3.23&0.86 &2.64 & 2.30& 5.71 \\
 Bridge & 2.34& 2.66& 55.63&57.38 &40.80 & 42.79&53.97 &55.28 \\
RailTrack &36.25 &55.81 &31.95 & 68.47&38.35 &59.94 & 41.26& 70.67\\
GroundRail &51.08 & 54.69&67.39 & 70.40& 66.54&68.71 & 74.91&80.92 \\
TrafficLight&49.71 & 63.47& 80.34& 85.57&78.50 &83.89 &75.89 & 79.93 \\
Static&24.07 & 26.66&32.90 & 36.36&38.18 &42.83 & 34.91& 37.69 \\
Dynamic &15.58 &18.24 & 24.06& 47.02&21.34 &55.09 & 35.15& 46.74\\
Water& 2.03&2.04 &8.66 & 8.67&18.25 &21.50 & 49.84& 56.63 \\
Terrain& 73.81& 87.82&75.50 &92.26 &78.37 & 92.15&76.70 & 92.91 \\
TwoWheeler & 55.65& 64.87&72.63 & 81.28&67.80 & 75.68& 66.65&77.25 \\
Bus & 74.87&80.73 &89.21 & 95.88&89.50 & 95.69&90.04 &94.73 \\
 Truck& 88.04&91.48 &92.45 &97.37 &92.10 &95.77 &91.94 & 94.54 \\

  \hline

\end{tabular}
}
\label{DELIVER_suppl}
\end{table*}

\begin{table*}[ht]
\centering
\caption{Class-Level semantic segmentation results on MCubes.}
\setlength{\tabcolsep}{12pt}
\renewcommand{\arraystretch}{1.4}
\resizebox{0.8\linewidth}{!}{
\begin{tabular}{ccccccc}
\hline
    \multirow{2}{*}{\textbf{Class}}     & \multicolumn{2}{c}{\textbf{I-A}} & \multicolumn{2}{c}{\textbf{I-A-D}} &  \multicolumn{2}{c}{\textbf{I-A-D-N}}   \\
    \cline{2-7} 
    &  \textbf{IoU} & \textbf{Acc} &  \textbf{IoU} & \textbf{Acc} &  \textbf{IoU} & \textbf{Acc}   \\ \hline
Asphalt & 79.83 & 98.06 & 85.93 & 96.19 & 89.22 & 97.31 \\
Concrete & 49.06 & 63.52 & 48.24 & 63.37 & 48.67 & 64.74 \\
Metal & 57.01 & 78.49 & 54.84 & 75.50 &53.63  & 78.16 \\
Road Marking & 72.55 & 86.51 & 74.49 & 83.03 & 74.36 & 81.35 \\
Fabric &38.76  & 44.96 & 35.38 & 39.32 & 28.97 & 33.58 \\
Glass & 61.93 & 77.48 & 60.78 & 80.74 & 60.95 & 77.31 \\
Plaster & 0.04 & 0.07 & 0.01 & 0.03 & 0.00 & 0.00 \\
Plastic & 20.59 & 23.13 & 26.27 & 31.13 & 22.53 & 26.02 \\
Rubber &30.53 &38.45 &32.75 &40.42 &31.94 & 40.13\\
Sand & 62.58 & 83.42 & 63.61 & 87.36 & 61.36 & 84.07 \\
Gravel & 4.49 & 4.84 & 51.50 & 55.28 & 59.40 & 68.50 \\
Ceramic & 36.93 & 46.43 & 35.19 & 43.80 & 29.98 & 35.90 \\
Cobblestone & 73.22 & 74.22 & 65.73 & 74.11 & 72.08 & 74.52 \\
Brick & 50.19 & 73.19 &44.33  & 66.68 & 46.94 & 65.31 \\
Glass & 62.01 & 78.47 & 61.86 & 74.86 & 58.38 & 74.57 \\
Wood & 51.42 & 65.85& 47.77 & 59.11 & 46.71 & 54.46 \\
Leaf & 76.51 & 87.67 & 76.76 & 88.65 & 77.47 & 89.03 \\
Water & 62.12 & 64.18 & 50.56 & 51.86 & 63.71 & 66.99 \\
Human & 37.69 & 46.47 & 31.25 & 40.37 & 34.54 & 39.73 \\
Sky & 96.60 & 98.46 & 96.65 & 98.56 & 96.65 & 98.37 \\

  \hline

\end{tabular}
}
\label{Mcubes_suppl}
\end{table*}
\end{document}